\documentclass[graybox]{svmult}

\usepackage{mathptmx}       % selects Times Roman as basic font
\usepackage{helvet}         % selects Helvetica as sans-serif font
\usepackage{courier}        % selects Courier as typewriter font
\usepackage{graphicx}        % standard LaTeX graphics tool
\usepackage{amsmath}
\usepackage{amssymb}

\DeclareMathOperator{\df}{def}

\makeindex             % used for the subject index
\begin{document}

\title*{A Stochastic Grammar for Natural Shapes}
\author{Pedro F. Felzenszwalb}
\institute{Pedro F. Felzenszwalb \at Brown University, \email{pff@brown.edu}}

\maketitle

\section*{Abstract}

We consider object detection using a generic model for natural
shapes. A common approach for object recognition involves matching
object models directly to images. Another approach involves building
intermediate representations via a generic grouping processes.  We
argue that these two processes (model-based recognition and grouping)
may use similar computational mechanisms.  By defining a generic model
for shapes we can use model-based techniques to implement a mid-level
vision grouping process.

\section{Introduction}

In this chapter we consider the problem of detecting objects using a
generic model for natural shapes.  A common approach for object
recognition involves matching object models directly to images.
Another approach involves building intermediate representations via
a generic grouping processes.  One of the ideas behind the work
described here is that these two processes (model-based recognition
and grouping) are not necessarily different.  By using a generic object
model we can use model-based techniques to perform
category-independent object detection.  This leads to a grouping
mechanism that is guided by a generic model for objects.

It is generally accepted that the shapes of natural objects have
certain regularities and that these regularities can be used to guide
visual perception.  For example, the Gestalt grouping laws explain how
the human visual system favors the perception of some objects over
others.  Intuitively, the tokens in an image should be grouped into
regular shapes because these groupings are more likely to correspond
to the actual objects in the scene.  This idea has been studied in
computer vision over several decades (see \cite{Ullman},
\cite{Nitzberg}, \cite{Mumford2}, \cite{Jacobs}, \cite{Lee},
\cite{Jermyn}).

We propose a method in which a generic process searches the image for
regular shapes to generate object hypotheses.  These hypotheses should
then be processed further in a way that depends on the perceptual task
at hand.  For example, each hypothesis could be matched against a
database of known objects to establish their identities.  Our
algorithm works by sampling shapes from a conditional distribution
defined by an input image.  The distribution is constructed so that
shapes with high probability look natural, and their boundaries align
with areas of the image that have high gradient magnitude.

Our method simply generates a number of potential object hypothesis.
Two hypothesis might overlap in the image, and some image areas might
not be in any hypothesis.  A consequence of this approach is that
the low-level processing doesn't commit to any particular
interpretation of the scene.

We start by defining a stochastic grammar that generates random
triangulated polygons.  This grammar can be tuned to capture regularities
of natural shapes.  For example, with certain choice of
parameters the random shapes generated tend to have piecewise smooth
boundaries and a natural decomposition into elongated parts.  We combine
this prior model with a likelihood model that defines the probability
of observing an image given the presence of a particular shape in the
scene.  This leads to a posterior distribution over shapes in a scene.
Samples from the posterior provide hypotheses for the objects in
an image.

Our approach is related to \cite{Zhu4} who also build a
stochastic model for natural shapes.  One important difference is that
our approach leads to polynomial time inference algorithms, while
\cite{Zhu4} relied on MCMC methods.

The ideas described here are based on the author's PhD thesis
\cite{thesis}.  

\section{Shape Grammar}

\begin{figure}
\centering
\includegraphics[height=1.5in]{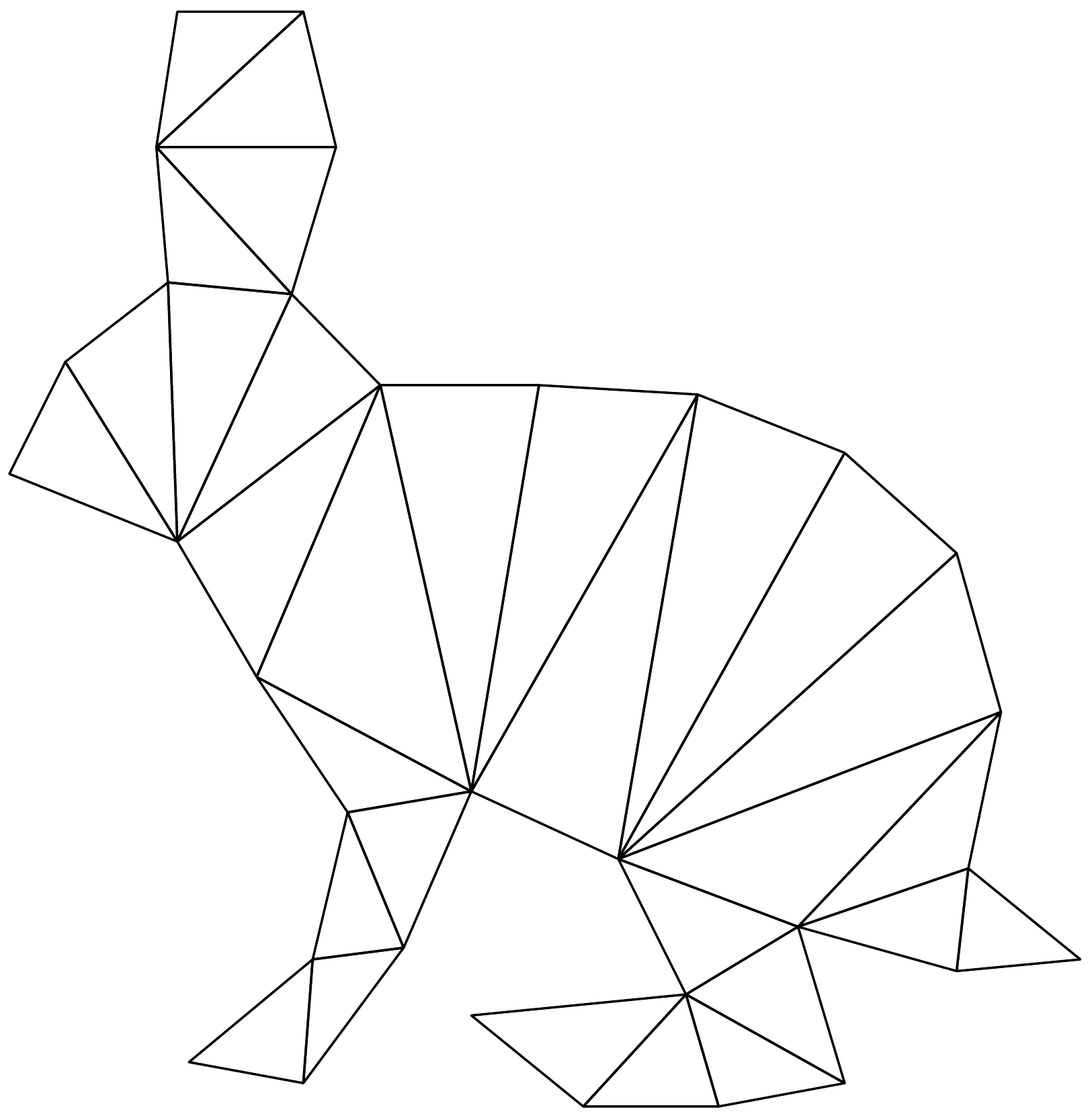}
\hspace{.3in}
\includegraphics[height=1.5in]{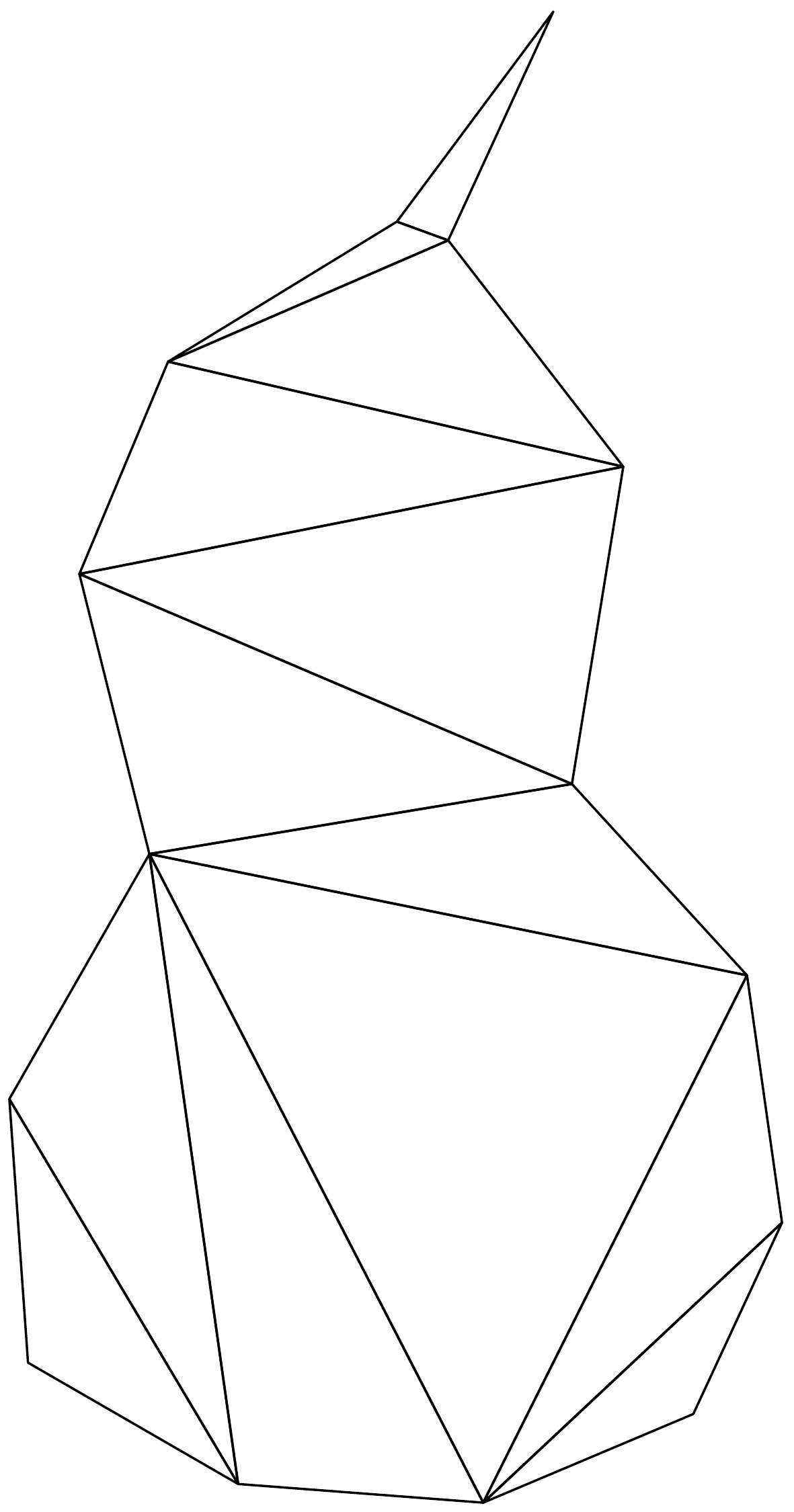}
\hspace{.3in}
\includegraphics[height=1.5in]{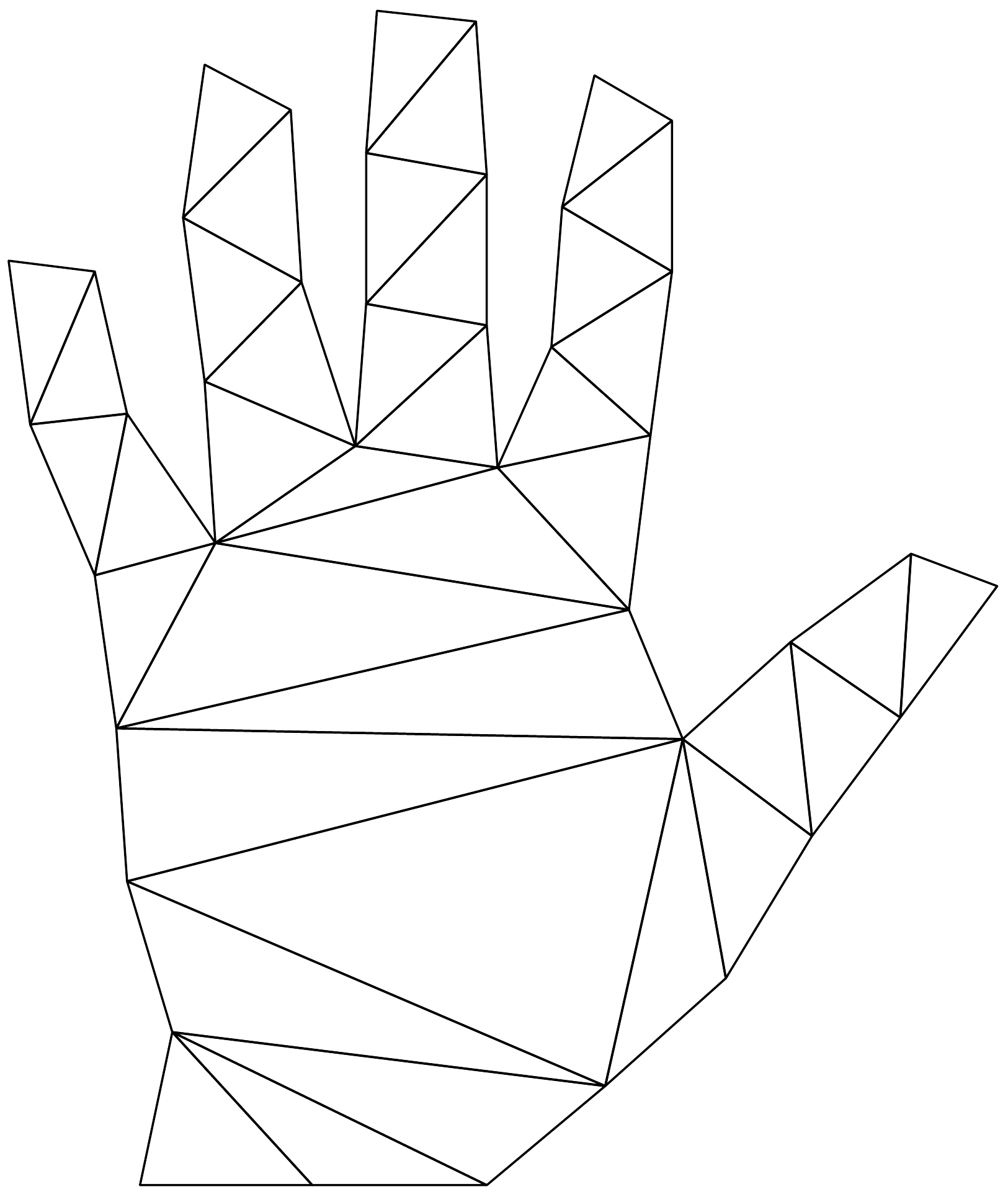}
\caption{Rabbit, pear, and hand represented by triangulated
polygons.  The polygonal boundaries represent the outlines, while the 
triangulations decompose the objects into parts.}
\label{fig:shapes}
\end{figure}

We represent objects using triangulated polygons.  Intuitively,
a polygonal curve is used to approximate the object boundary, and a
triangulation provides a decomposition of the objects into parts.
Some examples are shown in Figure~\ref{fig:shapes}.  

\begin{figure}
\centerline{\includegraphics[width=2.5in]{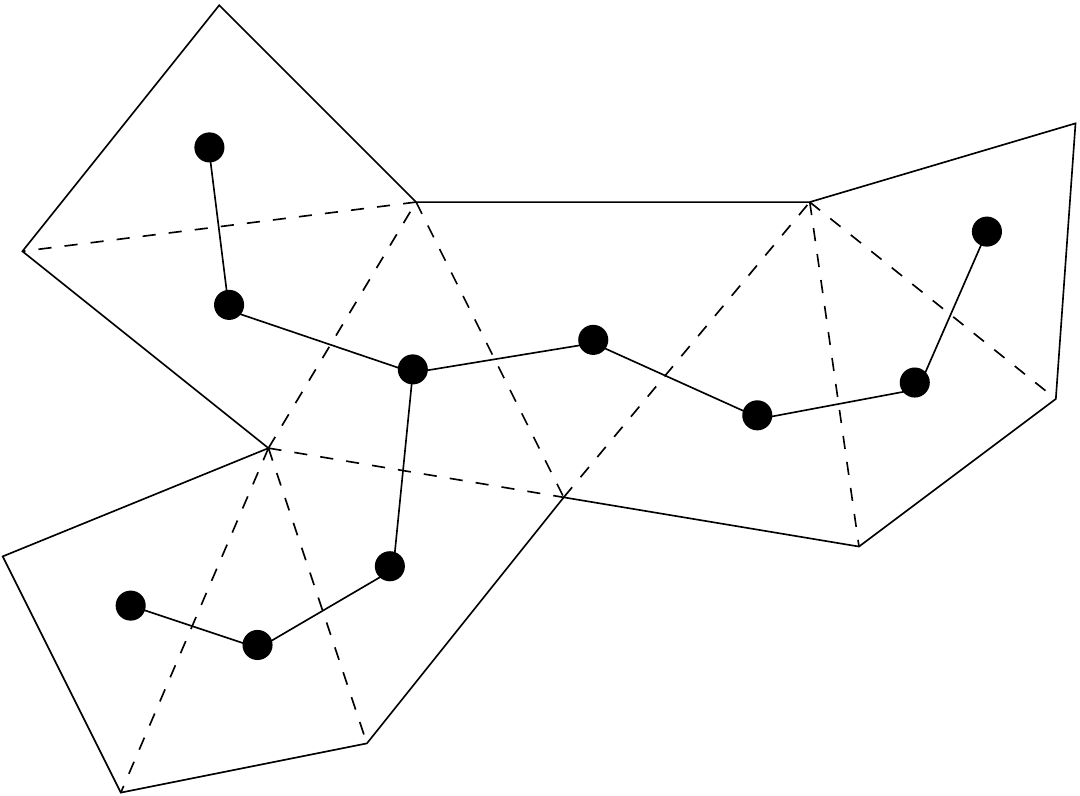}}
\caption{A triangulated polygon $T$ and its dual graph $G_T$.  If the polygon is
simple the dual graph is a tree where each node has degree 1, 2 or 3.}
\label{fig:polygon}
\end{figure}

There is a natural graph structure associated with a triangulated
polygon, where the nodes of the graph are the polygon vertices and the
edges include the polygon boundary and the diagonals in the
triangulation.  Figure~\ref{fig:polygon} shows a triangulated polygon
$T$ and its dual graph $G_T$.

\begin{figure}
\centerline{\includegraphics[width=4in]{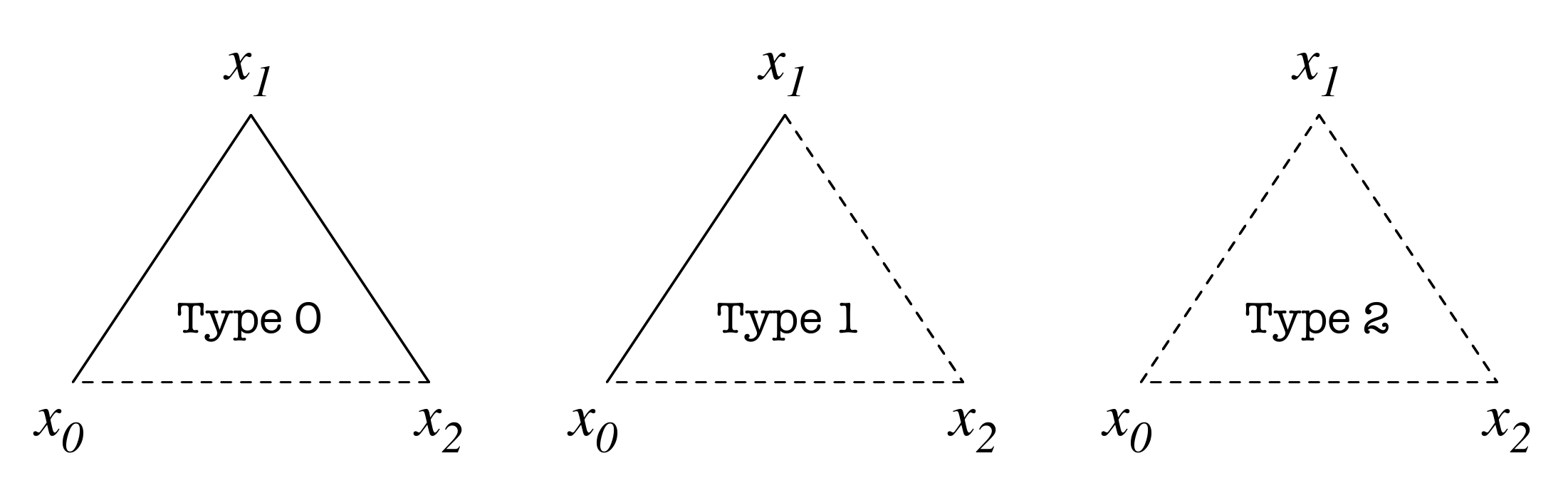}}
\caption{Different triangle types in a triangulated polygon.  The
  types corresponds to nodes of different degrees in the dual graph.
  Solid edges correspond to the polygon boundary while dashed edges
  are diagonals in the triangulation.}
\label{fig:triangles}
\end{figure}

Here we consider only objects that are represented by \emph{simple}
polygons (polygons without holes).  If $T$ is a triangulated simple
polygon, then its dual graph $G_T$ is a tree \cite{CG}.
There are three possible types of triangles in 
$T$, corresponding to nodes of different degrees in $G_T$.  The three
triangle types are shown in Figure~\ref{fig:triangles}, where solid
edges are part of the polygon boundary, and dashed edges are diagonals
in the triangulation.  Sequences of triangles of type 1 form branches, or
necks of a shape.  Triangles of the type 0 correspond to ends of
branches, and triangles of the type 2 form junctions connecting
multiple branches together.
For the rest of this chapter we will use a
particular labeling of the triangle vertices shown in
Figure~\ref{fig:triangles}.  A triangle will be defined by its
type (0,1 or 2) and the location of its vertices $x_0$, $x_1$ and $x_2$.

\begin{figure}
\centerline{\includegraphics[width=\linewidth]{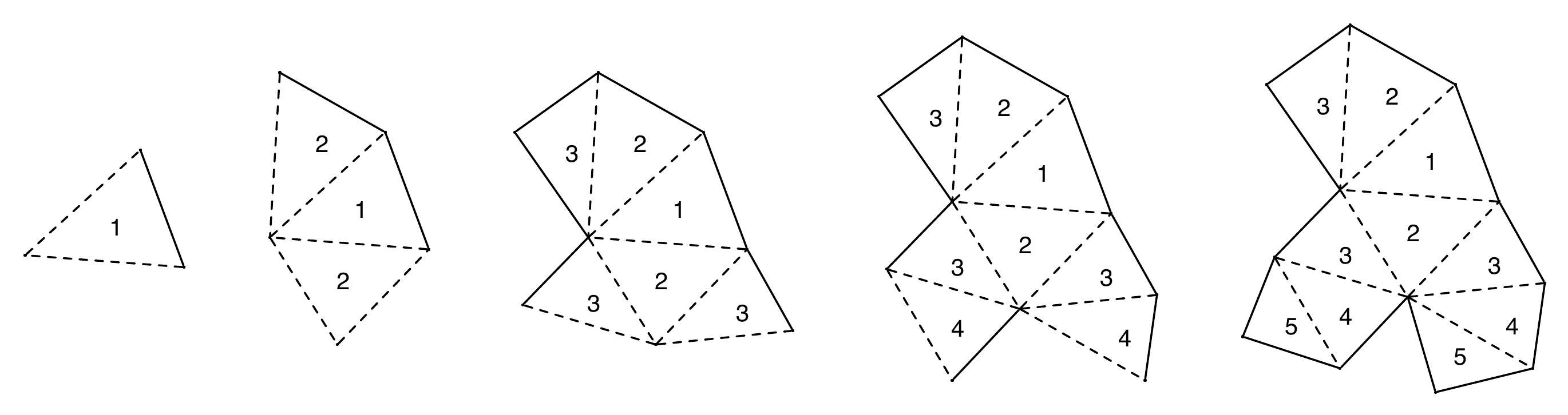}}
\caption{Growth of a triangulated polygon.  The label in each triangle
  indicates the stage at which it was created.  Initially we select a
  triangle (stage 1) from one of three possible types.  Then each
  dashed edge grows into a new triangle (stage 2) and growth continues
  along newly created dashed edges (stages 3, 4, 5).  New
  branches appear whenever a triangle of type 2 is created.  All
  branches end by growing a triangle of type 0.}
\label{fig:growth}
\end{figure}

A procedure to generate triangulated polygons is given by the
following growth process.  Initially a seed triangle is selected from
one of the three possible types.  Then each dashed edge ``grows'' into
a new triangle.  Growth continues along newly created dashed edges
until all branches end by growing a triangle of the first type.
Figure~\ref{fig:growth} illustrates the growth of a polygon.  A
similar process for growing combinatorial structures known as
$n$-clusters is described in \cite{Harary}.  The growth process can be
made stochastic as follows.  Let a triangle of type $i$ be selected
initially or during growth with probability $t_i$.  As an example,
imagine picking $t_i$ such that $t_1$ is large relative to $t_0$ and
$t_2$.  This would encourage growth of shapes with long branches.
Similarly, $t_2$ will control the number of branches in the shape.

The three parameters $t_0,t_1,t_2$ control the structure of the object
generated by the stochastic process.  The shape of the object is
determined by its structure and distributions that control the shape
of each triangle.  Let $X = (x_0, x_1, x_2)$ be the locations of the
vertices in a triangle.  We use $[X]$ to denote the equivalence class
of configurations that are equal up to translations, scales and
rotations.  The probability that a shape $[X]$ is selected for a
triangle of type $i$ is given by $s_i([X])$.  We assume the triangle
shapes are independent.\footnote{The fact that we can safely assume that
triangle shapes are independent in a triangulated polygon and get a sensible
model follows from Theorem 2.1 in \cite{thesis}.}  

\begin{figure}[t]
\centering
\includegraphics[width=2in]{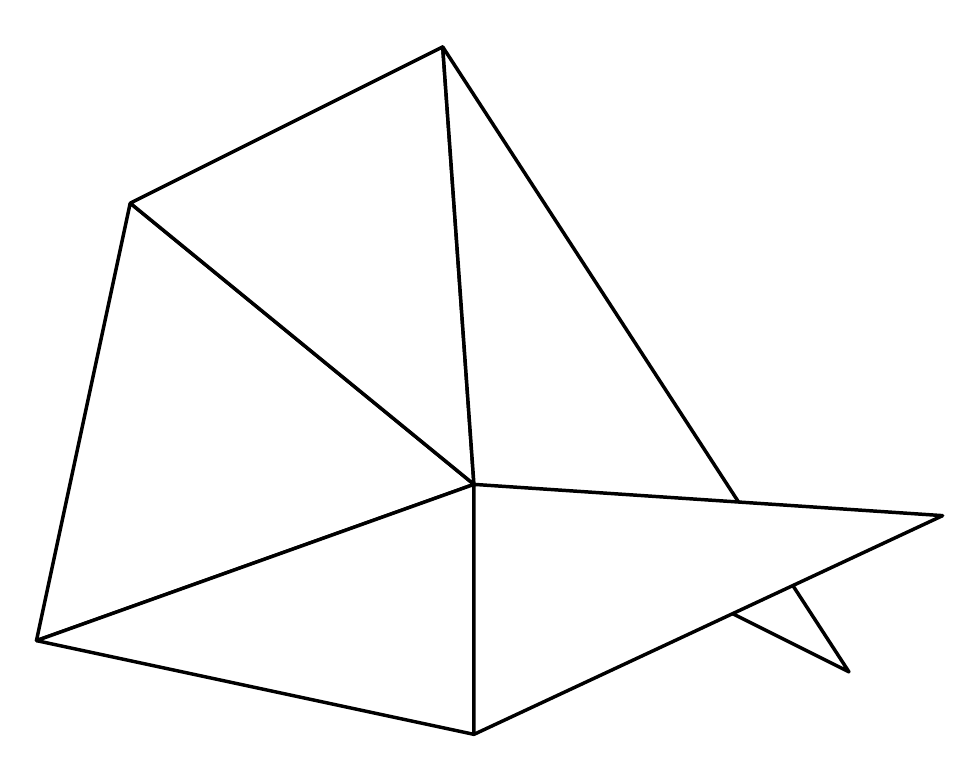}
\caption{In principle our growth process can generate objects with
  overlapping parts.}
\label{fig:crossing}
\end{figure}

The growth process described above can be characterized by a
stochastic grammar.  We note however that this grammar will not only
generate triangulated polygons, but will also generate objects with
overlapping parts as illustrated in Figure~\ref{fig:crossing}.

\begin{figure}
\centering
\includegraphics[width=\linewidth]{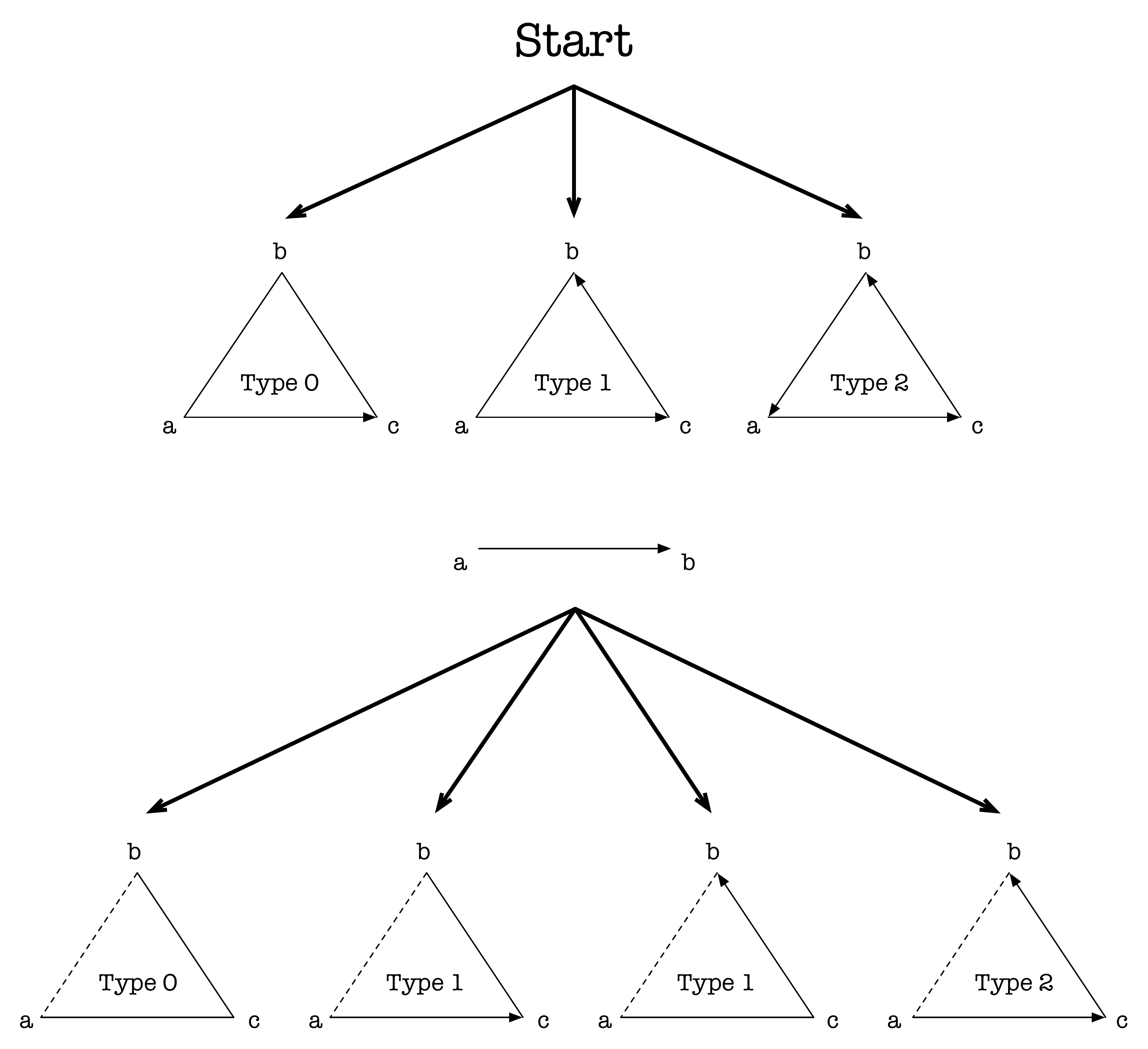}
\vspace{1cm}
\caption{Production rules for the shape grammar.  The grammar
  generates triangles and oriented edges.  The variables $a$, $b$ and
  $c$ correspond to locations in the plane.  The three variables are
  selected in a production from the start symbol, but only $c$ is
  selected in a production from an edge.  Note that edges are oriented
  carefully so that growth continues along a particular direction.}
\label{fig:grammar}
\end{figure}

There are two types of symbols in the grammar, corresponding to
triangles created during growth ${\cal T}$ and dashed edges that still
need to grow ${\cal E}$.  Triangles created during growth are elements
of ${\cal T} = \{0,1,2\} \times \mathbb{R}^2 \times \mathbb{R}^2
\times \mathbb{R}^2$.  The element $(i,a,b,c) \in {\cal T}$ specifies
a triangle of type $i$ with vertices $x_0=a$, $x_1=b$, $x_3=c$
following the labeling in Figure~\ref{fig:triangles}.  Edges that
still need to grow are elements of ${\cal E} = \mathbb{R}^2 \times
\mathbb{R}^2$.  The element $(a,b) \in {\cal E}$ specifies an internal
edge of the triangulated polygon from point $a$ to point $b$, The
edges are oriented from $a$ to $b$ 
so the system can ``remember'' the direction of
growth.  Figure~\ref{fig:grammar} illustrates the production rules for
the grammar.  Note that there are two different rules to grow a
triangle of type 1, corresponding to a choice of how the new triangle
is glued to the edge that is growing.  We simply let both choices have
equal probability, $t_1/2$.

%$$S \rightarrow B(a,b)B(b,c)E(c,a) \;\;|\;\; t_0 \times s_0([(a, b, c)])$$
%$$S \rightarrow B(a,b)E(b,c)E(c,a) \;\;|\;\; t_1 \times s_1([(a, b, c)])$$
%$$S \rightarrow E(a,b)E(b,c)E(c,a) \;\;|\;\; t_2 \times s_2([(a, b, c)])$$
%$$E(a,b) \rightarrow B(b,c)B(c,a) \;\;|\;\; t_0 \times s_0([(a, c, b)])$$
%$$E(a,b) \rightarrow B(b,c)E(c,a) \;\;|\;\; \frac{t_1}{2} \times s_1([(a, c, b)])$$
%$$E(a,b) \rightarrow E(b,c)B(c,a) \;\;|\;\; \frac{t_1}{2} \times s_1([(c, a, b)])$$
%$$E(a,b) \rightarrow E(b,c)E(c,a) \;\;|\;\; t_2 \times s_2([(a, b, c)])$$

To understand the effect of the parameters $t_0,t_1,t_2$, consider the
dual graph of a triangulated polygon generated by our stochastic
process.  The growth of the dual graph starts in a root node that has
one, two or three children with probability $t_0$, $t_1$ and $t_2$
respectively.  Now each child of the root grows according to a
Galton-Watson process \cite{Habib}, where each node has $i$ children
with probability $t_i$.

An important parameter of a Galton-Watson process is the expected
number of children for each node, or Malthusian parameter, that we
denote by $m$.  In our process, $m = t_1 + 2t_2$.  When $m < 1$ the
probability that the growth process eventually terminates is one.
From now on we will always assume that $m < 1$, which is equivalent to
requiring that $t_2 < t_0$ (here we use that $t_0+t_1+t_2=1$).  

%% In this case we can define a distribution over
%% triangulated polygons by letting $p(T)$ be proportional to the
%% probability that the grammar would generate $T$.  

Let $e$, $b$ and $j$ be random variables corresponding to the number
of end, branch and junction triangles in a random shape.  Let $n =
e+b+j$ be the total number of triangles in a shape.  For our
Galton-Watson process (corresponding to growth from each child of the
root of the dual graph) we can compute the expected number of nodes
generated, which we denote by $x$,
$$x = 1 + (x) t_1 + (2x) t_2 \; \Rightarrow \; x = 1/(t_0-t_2).$$ The
total number of triangles in a shape is obtained as one node for the
root of the dual graph plus the number of nodes in the subtrees
rooted at each child of the root.  So the expected value of $n$ is,
$$E(n) = 1 + (x) t_0+ (2x) t_1 + (3x) t_2.$$  Substituting for $x$
we get,
\begin{equation}
\label{eqn:size}
E(n) = \frac{2}{t_0 - t_2}.
\end{equation}

Similarly we can compute the expected value of $j$, the number of
junction triangles in a shape.  This quantity is
interesting because it gives a measure of the complexity of the shape.
In particular it is a measure of the number of parts (limbs, necks,
etc).  For the Galton-Watson process, let $y$ be the expected number
of nodes with degree 3 (two children),
$$y = (y) t_1 + (1+2y) t_2 \; \Rightarrow \; y = t_2/(t_0 - t_2).$$
The number of junction triangles in a shape equals the number of such
triangles in each subtree of the root plus one if the root itself
is a junction triangle,
$$E(j) = (y) t_0 + (2y) t_1 + (1+3y) t_2.$$  
Substituting for $y$ we get,
\begin{equation}
\label{eqn:junc}
E(j) = \frac{2t_2}{t_0 - t_2}.
\end{equation}

Equations~(\ref{eqn:size}) and~(\ref{eqn:junc}) provide intuition to the
effect of the parameters $t_0,t_1,t_2$.  The equations also show that
the parameters are uniquely defined by
the expected number of triangles and the expected number of junction
triangles in a random shape.  We can compute the $t_i$ corresponding
to any pair $E(n)$ and $E(j)$ such that $E(n) \geq 2$ and $E(n) \geq
2E(j)+2$.  These requirements are necessary since the
growth process always creates at least two triangles and the number of
triangles is always at least twice the number of junction triangles
plus two.

\begin{eqnarray*}
t_0 & = & (2+E(j))/E(n), \\
t_1 & = & 1-(2E(j)+2)/E(n), \\
t_2 & = & E(j)/E(n).
\end{eqnarray*}

\begin{figure}[t]
\centering
\includegraphics[width=\linewidth]{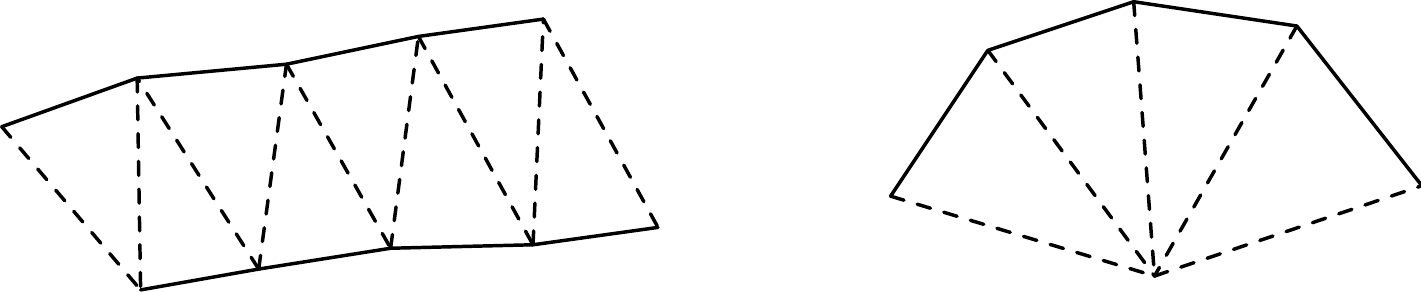}
\caption{Connecting multiple type 1 triangles in alternating
  orientations to form an elongated branch, and with the same
  orientation to form a bend.  If the neck triangles tend to be
  isosceles and thin than the shape boundary tends to be smooth.}
\label{fig:type1}
\end{figure}

While the $t_i$ control the combinatorial structure of the random
shapes we generate, their geometry is highly dependent on the choice
of shape for each triangle.  The triangle shapes are chosen according
to distributions that depend on the triangle type.  As an example we
can define,
$$s_i([X]) \propto e^{-k_i \df(X_i,X)^2},$$ where $X_i$ is an ideal
triangle of type $i$ and $\df(X_i,X)$ is the log-anisotropy of the
affine map taking $X_i$ to $X$ (see \cite{Dryden,thesis}).  The constant
$k_i$ controls how much the individual traingle shapes are allowed to
vary.  For the experiments in this chapter we chose both $X_0$ and
$X_2$ to be equilateral triangles and $X_1$ to be isosceles, with a
smaller side corresponding to the polygon boundary edge.  This choice
for $X_1$ generates shapes that tend to have smooth boundaries.
Figure~\ref{fig:type1} shows what happens when we connect multiple
triangles of this type with alternating or similar orientations.

\begin{figure}
\centering
\includegraphics[width=4.5in]{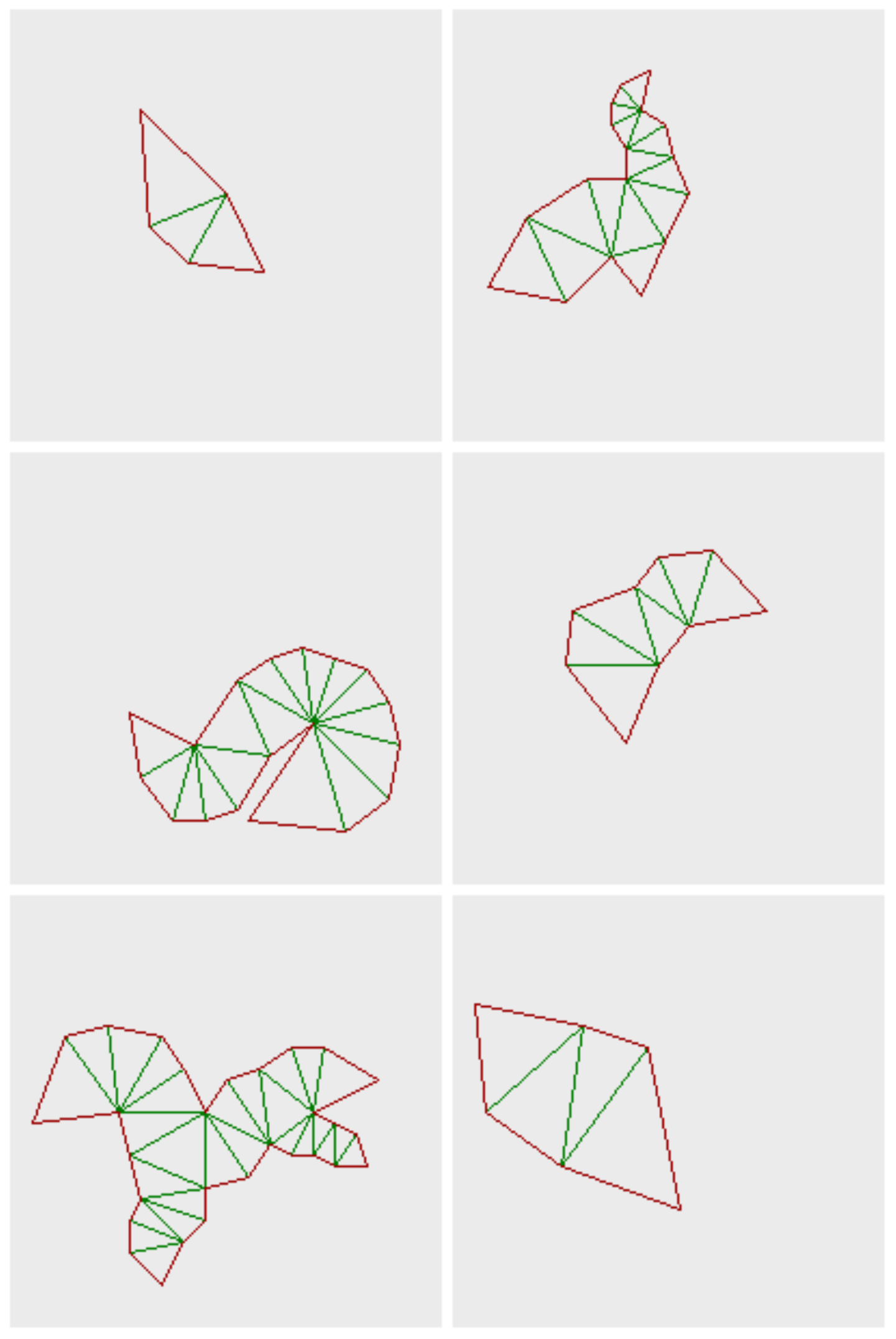}
\caption{Examples of random shapes generated by the stochastic grammar.}
\label{fig:random}
\end{figure}

Figure~\ref{fig:random} shows some random shapes generated by the
random process with $E(n) = 20$, $E(j) = 1$, and the choice for
$s_i([X])$ described above.  Note how the shapes have natural
decompositions into parts, and each part has an elongated structure,
with smooth boundaries almost everywhere.  These examples illustrate
some of the regularties captured by our stochastic shape grammar.  In
the next section we will show how the grammar can be used for
object detection.

\section{Sampling Shapes From Images}

Now we describe how our model for random shapes can be combined with a
likelihood function to yield a posterior distribution $p(T|I)$ over
triangulated polygons in an image.  We then show how to sample from
the posterior using a dynamic programming procedure.  The approach is
similar to sampling from the posterior distribution of a hidden Markov
model using weights computed by the forward-backward algorithm
\cite{Rabiner}.  Our experiments in the next Section illustrate how
samples from $p(T|I)$ provide hypotheses for the objects in an
image.

Recall that each triangle created during growth is an element of ${\cal T}$,
specifying a triangle type and the location of its vertices.  We
assume that the likelihood $p(I|T)$ factors into a product of terms,
with one term for each triangle,
\begin{equation}
p(I|T) \propto \prod_{(i,x_0,x_1,x_2) \in T} \pi_i(x_0,x_1,x_2,I). 
\label{eqn:likelihood}
\end{equation}
This factorization allows for an efficient inference algorithm to
be developed to generate samples from the posterior $p(T|I) \propto p(I|T)p(T)$.

We expect the image to have high gradient at the boundary of objects,
with orientation perpendicular to the boundary.
In practice we have used a likelyhood function of the form,
$$P(I|T) \propto \exp\left(\lambda \int \Vert (\nabla I \circ f)(s)
\times f'(s) \Vert \; ds\right).$$ Here $f(s)$ is a parametrization of
the boundary of $T$ by arclength.  The term $\Vert (\nabla I \circ
f)(s) \times f'(s) \Vert$ is the component of the image gradient that
is perpendicular to the object boundary at $f(s)$.  The integral above
can be broken up into a sum of terms, with one term for each boundary
edge in the triangulated polygon.  This allows us to write the
likelihood in the form of equation~(\ref{eqn:likelihood}) where
$\pi_i(x_0,x_1,x_2,I)$ evaluates the contribution to the integral due
to the boundary terms (solid edges) of a triangle of type $i$ with
vertices $(x_0,x_1,x_2)$.

Let $T_r$ denote a triangulated polygon rooted at a triangle $r$.  Using
Bayes' law we can write the posterior distribution for rooted shapes
given an observed image as,
$$p(T_r|I) \propto p(T_r) p(I|T).$$ There are two approximations we
make to sample from this posterior efficiently.  We
consider only shapes where the depth of the dual graph is bounded by a
constant $d$ (the depth of a rooted graph is the maximum distance from
a leaf to the root).  This should not be a significant problem since
shapes with too many triangles have low prior probability anyway.
Moreover, the running time of our sampling algorithm is linear in $d$,
so we can let this constant be relatively large.  We also only
consider shapes where the location of each vertex is constrained to
lie on a finite grid ${\cal G}$, as opposed to an arbitrary location
in the plane.  The running time of our algorithm for sampling
from $p(T|I)$ is $O(d|{\cal G}|^3)$.

To sample from the posterior we first pick a root triangle, then pick
the triangles connected to the root and so on.  The root triangle $r$ should be
selected according to its marginal conditional distribution,
\begin{equation}
p(r|I) = \sum_{T_r} p(T_r|I).
\label{eqn:marg}
\end{equation} 
Note that the sum is over all shapes rooted at $r$, and with the depth
of the dual graph bounded by $d$.  We can compute this marginal
distribution in polynomial time because the triangles in a shape are
connected together in a tree structure.

\begin{figure}[t]
\centering
\includegraphics[width=1.8in]{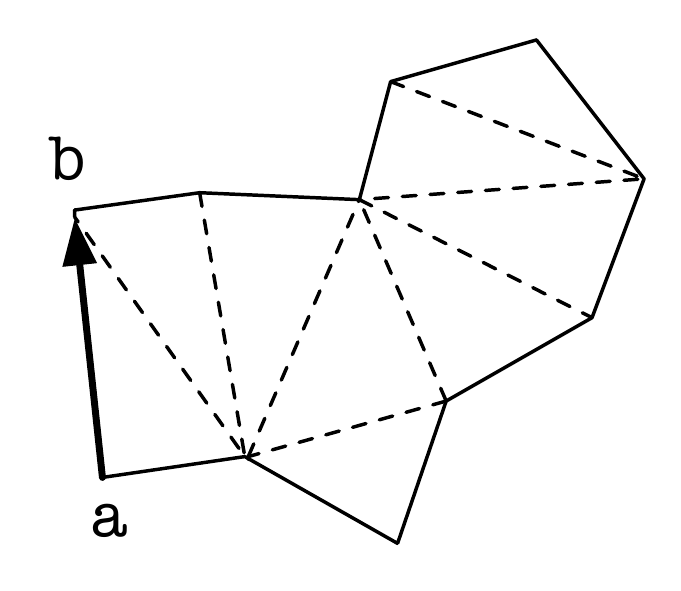}
\caption{A partial shape generated from the
edge $(a,b)$.}
\label{fig:partial}
\end{figure}

Let $T_{(a,b)}$ denote a partial shape generated from an edge $(a,b)$.
Figure~\ref{fig:partial} shows an example of a partial shape.  We
denote the probability that the grammar would generate $T_{(a,b)}$
starting from the edge $(a,b)$ by $p(T_{(a,b)})$.  The posterior
probability of a partial shape $T_{(a,b)}$ given an image $I$ is given
by,
$$p(T_{(a,b)}|I) \propto p(T_{(a,b)}) 
\prod_{(i,x_0,x_1,x_2) \in T_{(a,b)}} \pi_i(x_0,x_1,x_2,I).$$ 

We define the following quantities in analogy to the 
backward weights of a hidden Markov nodel (see \cite{Rabiner}),
$$V_j(a,b) = \sum_{T_{(a,b)}} p(T_{(a,b)}|I),$$ where the sum is taken
over all partial shapes with depth at most $j$.  Here we measure
depth by imagining the root to be a triangle that would be immediately
before the edge $(a,b)$.  The quantities $V_j(a,b)$ can be computed
recursively using a dynamic programming procedure,
\begin{eqnarray*}
V_0(a,b) & = & 0, \\
V_j(a,b) & = & t_0 \sum_{c} \, s_0([b,c,a]) \, \pi_0(b,c,a,I) + \\
& & (t_1/2) \sum_{c} \, s_1([b,c,a]) \, \pi_1(b,c,a,I) \, 
V_{j-1}(a,c) + \\
& & (t_1/2) \sum_{c} \, s_1([c,a,b]) \, \pi_1(c,a,b,I) \, 
V_{j-1}(c,b) + \\
& & t_2 \sum_{c} \, s_2([b,c,a]) \, \pi_2(b,c,a,I) \, 
V_{j-1}(a,c) \, V_{i-1}(c,b).
\end{eqnarray*}
Now, depending on the type of the root triangle
we can rewrite the marginal distribution in equation~(\ref{eqn:marg})
as,
\begin{eqnarray*}
p((0,a,b,c)|I) & \propto & t_0 \, s_0([a,b,c]) \, V_d(a,c), \\
p((1,a,b,c)|I) & \propto & t_1 \, s_1([a,b,c]) \, V_d(a,c) \, V_d(c,b), \\
p((2,a,b,c)|I) & \propto & t_2 \, s_2([a,b,c]) \, V_d(a,c) \, V_d(c,b) \, 
V_d(b,a).
\end{eqnarray*}
The equations above provide a way to sample the root triangle from its
marginal distribution.  The running time for computing all
the $V_j(a,b)$ and the marginal distribution for the root triangle is
$O(d|{\cal G}|^3)$.  Once we compute these quantities we can obtain
samples for the root by sampling from a discrete
distribution.  After choosing $r = (i,x_0,x_1,x_2)$ we need to sample the
triangles connected to the root.  We then sample the triangles that
are at distance two from the root, and so on.  When sampling a
triangle at distance $j$ from the root, we have an edge $(a,b)$ that
is growing.  We need to sample a triangle by selecting the location ${\bf c}$ of a new vertex and
a triangle type according to 
\begin{eqnarray*}
p((0,b,{\bf c},a)|I,(a,b)) & \propto & t_0 \, s_0([b,{\bf c},a]), \\
p((1,b,{\bf c},a)|I,(a,b)) & \propto & (t_1/2) \, s_1([b,{\bf c},a]) 
\, V_{d-j}(a,{\bf c}), \\
p((1,{\bf c},a,b)|I,(a,b)) & \propto & (t_1/2) \, s_1([{\bf c},a,b]) 
\, V_{d-j}({\bf c},b), \\
p((2,b,{\bf c},a)|I,(a,b)) & \propto & t_2 \, s_2([b,{\bf c},a]) \, V_{d-j}(a,{\bf c}) \, 
V_{d-j}({\bf c},b).
\end{eqnarray*}
We evaluate these probabilities using the precomputed $V_j$ quantities
and then sample a triangle type and location $c$ from the
corresponding discrete distribution.  Note that for a triangle at
depth $d$ the only choices with non-zero probability will have type
zero, as $V_0(a,b) = 0$.

\section{Experimental Results}

For the experiments in this section we used a grid ${\cal G}$ of $40
\times 40$ locations for the vertices of the shapes.  We used the
likelihood model defined in the last section, and the same grammar
parameters used to generate the random shapes in
Figure~\ref{fig:random}.

\begin{figure}
\centering
\includegraphics[width=\linewidth]{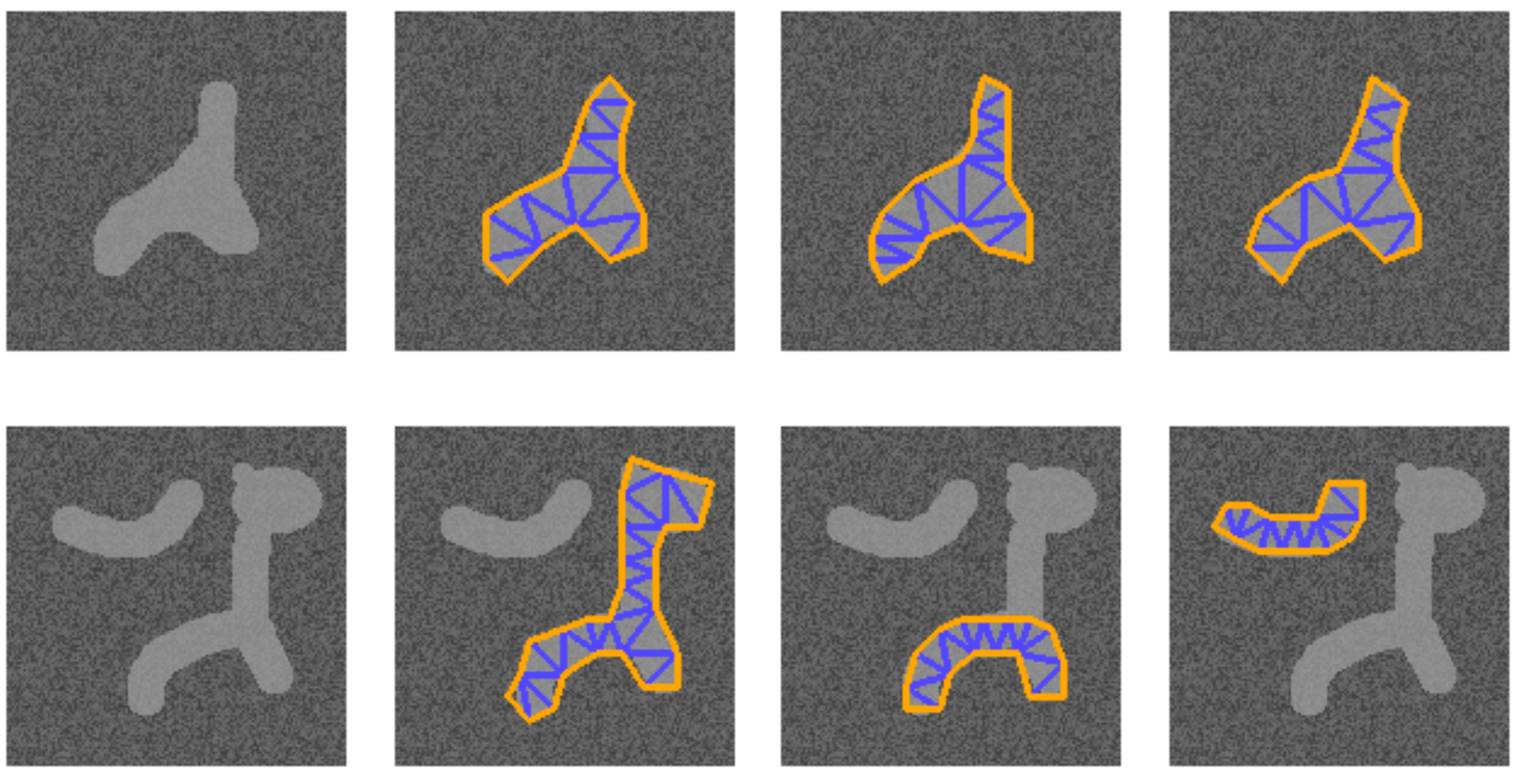} 
\caption{Samples from $p(T|I)$ for two synthetic images $I$.  Note how in the
second image we get multiple potential objects among the samples.}
\label{fig:synt}
\end{figure}

Figure~\ref{fig:synt} shows some of the samples generated from the
posterior distribution $p(T|I)$ for two different synthetic images.
The first image has a single object and each sample from
$p(T|I)$ gives a slightly different representation for that object.
The second image has two objects and the
samples from $p(T|I)$ are split between the two objects.  Note that we
obtain samples that correspond to each object and also to a part of
one object that can be naturally interpreted as a single object.
Overall the samples in both cases give resonable interpretations of
the objects in the images.

\begin{figure}
\centering
\begin{tabular}{cc}
\includegraphics[width=1.5in]{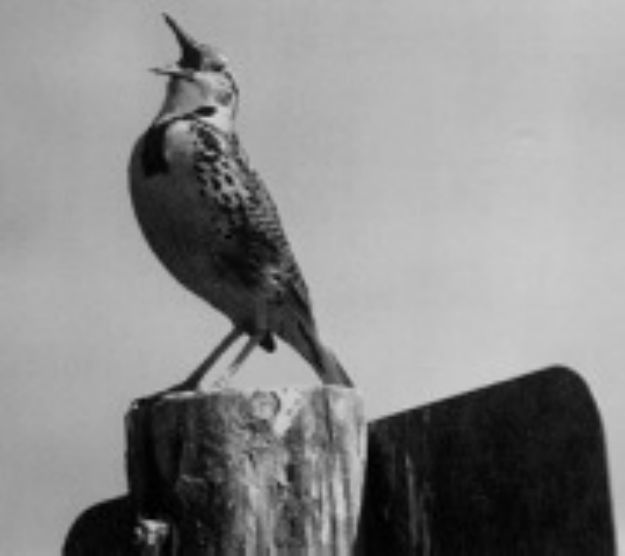} &
\includegraphics[width=1.5in]{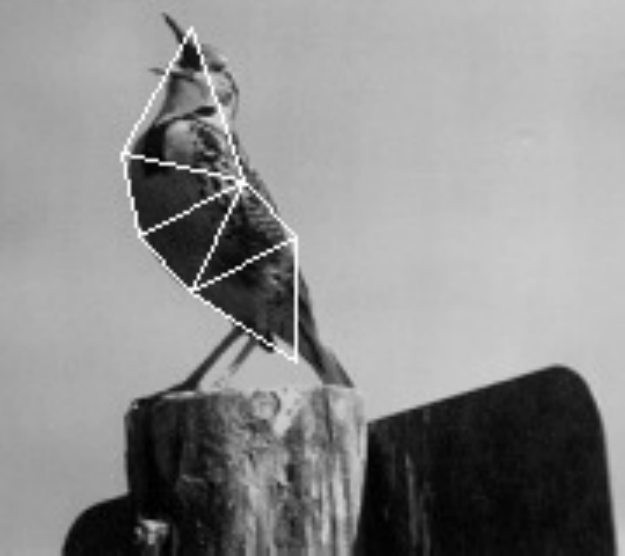} 
\includegraphics[width=1.5in]{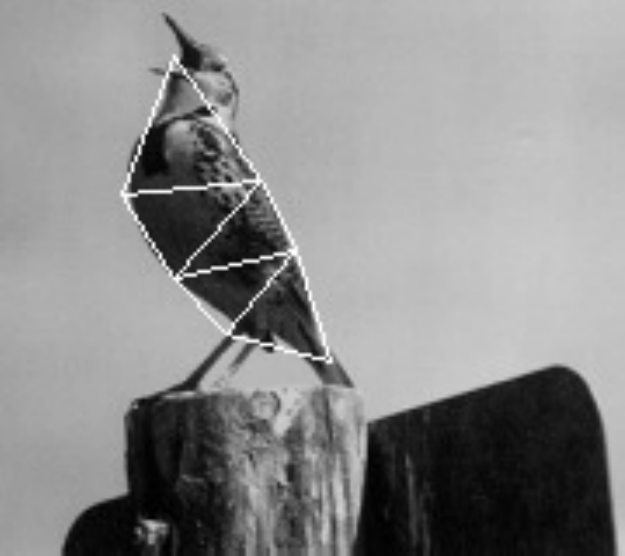}
\end{tabular}
\caption{Sample from $p(T|I)$ for an image with a bird.}
\label{fig:bird}
\end{figure}

\begin{figure}
\centering
\begin{tabular}{ccc}
\includegraphics[width=1.45in]{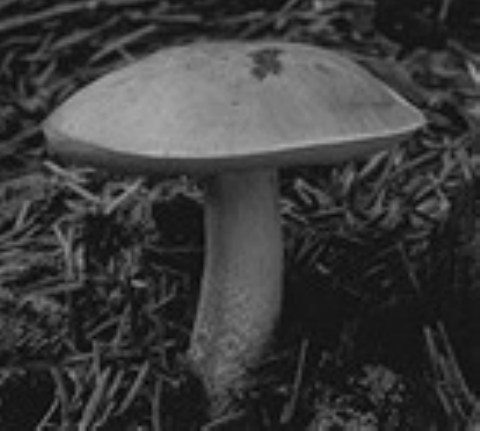} &
\includegraphics[width=1.45in]{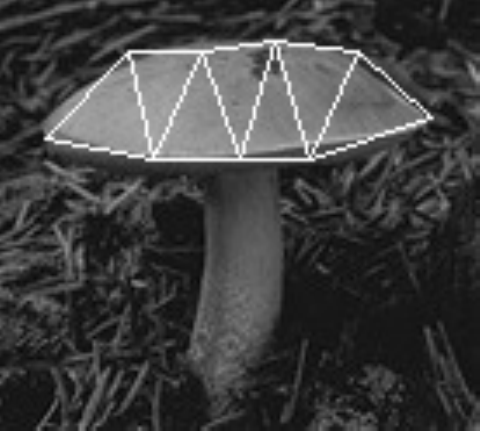} &
\includegraphics[width=1.45in]{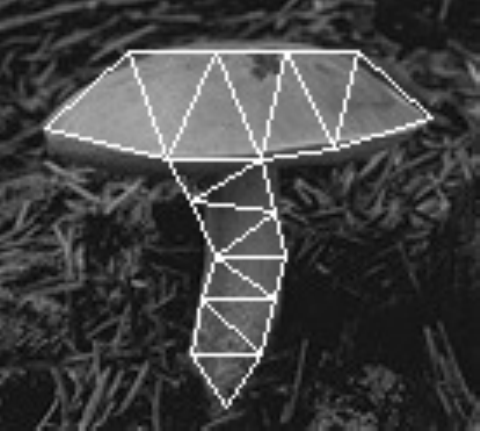}
\end{tabular}
\caption{Samples from $p(T|I)$ for an image with a mushroom.}
\label{fig:mush}
\end{figure}

Figures~\ref{fig:bird} and~\ref{fig:mush} show samples from the posterior
distributon $p(T|I)$ for two natural images.  In practice we obtain groups
of samples that are only slightly different from each other, and here we show
representatives from each group.
For the mushroom image, we obtained different samples
corresponding to competing interpretations.  In one case
the whole mushroom is considered as an object, while in another case the
stem comes out on its own.

\newpage

\bibliographystyle{plain}
\bibliography{sgrammar}

\begin{thebibliography}{10}

\bibitem{CG}
Mark De~Berg, Otfried Cheong, Marc Van~Kreveld, and Mark Overmars.
\newblock {\em Computational geometry: algorithms and applications}.
\newblock Springer, 2008.

\bibitem{Dryden}
I.L. Dryden and K.V. Mardia.
\newblock {\em Statistical Shape Analysis}.
\newblock John Wiley {\&} Sons, 1998.

\bibitem{thesis}
P.~F. Felzenszwalb.
\newblock {\em Representation and Detection of Shapes in Images}.
\newblock PhD thesis, MIT, September 2003.

\bibitem{Habib}
M.~Habib, C.~McDiarmid, J.~Ramirez-Alfonsin, and B.~Reed.
\newblock {\em Probabilistic Methods for Algorithmic Discrete Mathematics}.
\newblock Springer-Verlag, 1998.

\bibitem{Harary}
F.~Harary, Palmer E.M., and R.C. Read.
\newblock On the cell-growth problem for arbitrary polygons.
\newblock {\em Discrete Mathematics}, 11:371--389, 1975.

\bibitem{Jacobs}
D.W. Jacobs.
\newblock Robust and efficient detection of salient convex groups.
\newblock {\em IEEE Transactions on Pattern Analysis and Machine Intelligence},
  18(1):23--37, 1996.

\bibitem{Jermyn}
I.H. Jermyn and H.~Ishikawa.
\newblock Globally optimal regions and boundaries as minimum ratio weight
  cycles.
\newblock {\em IEEE Transactions on Pattern Analysis and Machine Intelligence},
  23(10):1075--1088, October 2001.

\bibitem{Lee}
M.S. Lee and G.~Medioni.
\newblock Grouping ., -, -->, 0, into regions, curves, and junctions.
\newblock {\em Computer Vision and Image Understanding}, 76(1):54--69, 1999.

\bibitem{Mumford2}
D.~Mumford.
\newblock Elastica and computer vision.
\newblock In {\em Algebraic Geometry and Its applications}, pages 491--506.
  Springer-Verlag, 1994.

\bibitem{Nitzberg}
M.~Nitzberg and D.~Mumford.
\newblock The 2.1-d sketch.
\newblock In {\em ICCV}, pages 138--144, 1990.

\bibitem{Rabiner}
L.~R. Rabiner.
\newblock A tutorial on hidden markov models and selected applications in
  speech recognition.
\newblock {\em Proceedings of the IEEE}, 77(2), February 1989.

\bibitem{Ullman}
A.~Shashua and S.~Ullman.
\newblock Structural saliency: The detection of globally salient structures
  using a locally connected network.
\newblock In {\em ICCV}, pages 321--327, 1988.

\bibitem{Zhu4}
S.C. Zhu.
\newblock Embedding gestalt laws in markov random fields.
\newblock {\em IEEE Transactions on Pattern Analysis and Machine Intelligence},
  21(11):1170--1187, 1999.

\end{thebibliography}

\end{document}